\pdfoutput=1
\documentclass[11pt]{article}

\usepackage[]{emnlp2021}

\usepackage{times}
\usepackage{latexsym}

\usepackage[T1]{fontenc}
\usepackage[utf8]{inputenc}

\usepackage{microtype}
\usepackage{multirow}
\usepackage{threeparttable}
\usepackage{bm}
\usepackage{amsfonts}
\usepackage{listings}
\usepackage{booktabs}
\usepackage{amsmath}
\usepackage{graphicx}
\usepackage{subcaption}
\usepackage{enumitem}
\setenumerate[1]{itemsep=0pt,partopsep=0pt,parsep=\parskip,topsep=5pt}
\setitemize[1]{itemsep=0pt,partopsep=0pt,parsep=\parskip,topsep=5pt}
\setdescription{itemsep=0pt,partopsep=0pt,parsep=\parskip,topsep=5pt}
\title{Topic-Guided Abstractive Text Summarization: a Joint Learning Approach}

\author{Chujie Zheng{$^{1,*}$}, Kunpeng Zhang{$^{2}$}, Harry Jiannan Wang{$^{1}$}, Ling Fan{$^{3,4}$}, Zhe Wang{$^{4}$} \\
        {$^{1}$}University of Delaware \\ {$^{2}$}University of Maryland, College Park \\ {$^{3}$}Tongji University, {$^{4}$}Tezign \\
        \texttt{chz@udel.edu}, \texttt{kpzhang@umd.edu}, \texttt{hjwang@udel.edu},\\\texttt{lfan@tongji.edu.cn}, \texttt{w@tezign.com}}

\date{}

\begin{document}
\maketitle
\begin{abstract}
We introduce a new approach for abstractive text summarization, Topic-Guided Abstractive Summarization, which calibrates long-range dependencies from topic-level features with globally salient content. The idea is to incorporate neural topic modeling with a Transformer-based sequence-to-sequence (seq2seq) model in a joint learning framework. This design can learn and preserve the global semantics of the document, which can provide additional contextual guidance for capturing important ideas of the document, thereby enhancing the generation of summary. We conduct extensive experiments on two datasets and the results show that our proposed model outperforms many extractive and abstractive systems in terms of both ROUGE measurements and human evaluation.
\end{abstract}

\section{Introduction}
Automatic summarization has made impressive success with the advance of large-scale language models. As one of the central problems in Natural Language Processing (NLP), summarization aims at generating an accurate text snippet to capture the key information of the input text. Comparing to extractive summarization which copies informative fragments from the input, abstractive summarization requires understanding the input document at first and subsequently generate a summary with novel and relevant words. A good abstractive summary is expected to cover principal information in the input, as well as being linguistically fluent.

In abstractive summarization, sequence-to-sequence (Seq2Seq) \cite{sutskever2014sequence} has become a dominant framework using an encoder-decoder architecture based on RNNs \cite{chung2014empirical} and more recently Transformers \cite{vaswani2017attention}. %Most prior work on neural abstractive summarization relied on high-quality datasets of supervised document-summary pairs. Within the success of large-scaled self-supervised language models \cite{devlin2019bert, peters2018deep, radford2019language, liu2019roberta}, 
This framework has often been utilized for language generation tasks, where two typical steps are involved: (1) defining the self-supervised pretraining objective for Transformer-based sequence-to-sequence models, and (2) fine-tuning it on the datasets for the downstream tasks. This framework has extended the success of text generation to summarization and achieved many promising results. Current state-of-the-art (SOTA) summarization models, including BART \cite{lewis2020bart}, PEGASUS \cite{zhang2020PEGASUS} and ProphetNet \cite{qi2020prophetnet}, all adopted this Transformer-based architecture. Most transformer-based models are better at exploring the relationships among local tokens \cite{wang2020friendly}. However, the performance of semantic understanding at a higher level (e.g, sentences, topics) is usually subpar \cite{reimers2019sentence}.

To better capture the global semantics of an input document for language generation tasks, researchers have made many attempts to improve current models. For example, topic-aware methods have been proposed to assist document summarization \cite{wang2018reinforced, ailem2019topic, narayan2018don, wang2020friendly, fu2020document}. Topic models, such as LDA \cite{blei2003latent}, PFA \cite{zhou2012beta} and prodLDA \cite{srivastava2017autoencoding}, are able to provide additional signals for document understanding \cite{peinelt2020tbert}. They all consider topics as global variables to describe the distribution over all tokens in the vocabulary \cite{wang2020friendly}. The topic is proven to be especially useful for dealing with domain-specific language since topic models have been exploited for domain adaption \cite{hu2014polylingual, guo2009domain}. For text summarization, by incorporating the topic-level features into the summarization model, we believe it can improve model performance since it encourages the model to focus on both local relationships and global semantics.

Despite the amount of literature in combing topic modeling into NLP tasks, we find that many prior studies use the topic model as a separate component for information extraction rather than jointly improve NLP tasks and topic modeling in a unified way. For example \citet{bianchi2020pre, bianchi2020cross} developed a topic model where the input of an document is not traditional bag-of-words but representation learned via BERT \cite{devlin2019bert}. Most cutting-edge topic models adopt the encoder-decoder network architecture where an input document is encoded into latent topic factors (i.e., coding) and these coding is then used to reconstruct the original input document during the decoding process. Such a design is very similar to seq2seq architecture where the input is also first converted to latent variables as input for decoder. Therefore, we believe embedding topic modeling into a seq2seq-based summarization model can improve performance. 

In this paper, we design a topic-guided language generation framework for Transformer-based language models, by understanding the semantics learned via the topic model. We incorporate the topic model into a joint learning framework with the language generation and use the information provided by the topic model to guide the language generation. To demonstrate the effectiveness of our proposed method, we conduct extensive experiments on two real-world datasets. Both the quantitative results and the human evaluation feedback confirm that our model is more capable of capturing the key information for abstractive text summarization. Our key contributions are three-fold:
\begin{itemize}[leftmargin=*]
    \item We propose a new framework for abstractive text summarization by incorporating topic modeling in a joint learning manner, which helps capture the global topic semantics for better summarization generation. 
    % This generic framework opens a new perspective in NLP and can be extended to other language tasks.
    \item We design a joint learning architecture to fuse neural topic modeling with transformer-based  sequence-to-sequence model.
    \item We evaluate our topic-guided abstractive summarization model on two standard summarization datasets and demonstrate performance improvements over existing methods in both quantitative measurement and human evaluation. 
\end{itemize}

\section{Related Work}
We now review the related literature grouped in three main categories and position our work in that context by indicating the respective issues (that are addressed by our contributions).

\textbf{Abstractive text summarization} has been widely explored over the past decades. Pre-trained encoder-decoder Transformers-based models have exhibited great success and become the first choice for many NLP tasks due to their long-term dependency modeling capability and scalability, particularly for text summarization tasks \cite{lewis2020bart, zhang2020PEGASUS, qi2020prophetnet}. Despite their outstanding performance on standard quantitative measurement, a common criticizing point is that higher-level global semantic structure in the text is usually ignored. For the summarization task, this can lead to the significantly inferior performance \cite{reimers2019sentence}.

\textbf{Topic model} explores the hidden semantic structures of text \cite{wang2020learning}. One basic assumption for the topic model is that a document is a mixture of topics and each topic is distributed over words in the vocabulary of the corpus. To learn these distributions, Latent Dirichlet Allocation (LDA) is introduced by imposing latent variables with Dirichlet prior \cite{griffiths2004hierarchical}. Recently, given the power of neural networks, topic models are improved and implemented through generic auto-encoder \cite{larochelle2012neural, salakhutdinov2009replicated} or neural variational auto-encoder \cite{miao2016neural, miao2017discovering, roder2015exploring, srivastava2017autoencoding, ding2018coherence}. 
% \citet{miao2016neural} propose an unsupervised generative model NVDM based on a VAE, assuming a Gaussian distribution over topics. \citet{srivastava2017autoencoding} propose a neural variational framework that explicitly approximates the Dirichlet prior, using a Gaussian distribution to obtain more interpretable and coherent topics. \citet{miao2017discovering} parameterize the multinomial distributions of each document, proposing three variants of neural topic models to exhibit sparse topic distributions.

Topic model has been incorporated to different types of NLP tasks, including document classification \cite{wang2020learning}, translation \cite{zhang2016topic}, and summarization \cite{ailem2019topic, narayan2018don, wang2020friendly, fu2020document}. \citet{zhang2016topic} build a topic-informed RNN considering each word as a distribution over topics for machine translation. \citet{ailem2019topic} develop a topic augmented decoder that generates a summary conditioned on both the input document and the latent topic for the document. \citet{narayan2018don} propose the topic-conditioned Seq2Seq model under the CNN framework. \citet{wang2020friendly} introduce the friendly topic assistant for Transformer-based summarization models with the topic information explored by the separate topic component. Using the topic model to capture the global semantics has been proven to be an effective means for many NLP tasks. The topic model can be used as a separate component to provide additional information that can be used as an extra feature \cite{wang2020friendly, zou2020topic, fu2020document}. This topic-level feature has been applied to the language generation process \cite{ailem2019topic, wang2019topic}, to guide text generation with designated topic guidance.

\section{Methodology}

\subsection{Transformer-based Seq2Seq Model}
\label{section::transformer-based-seq2seq-model}
Natural language generation tasks are best expressed as sequence-to-sequence (Seq2Seq) problems. They often assume that each word is encoded into a vector representation. Then a document $\textbf{d}$ with $n$ words can be represented as a sequence of $n$ vectors: $\textbf{d}=\textbf{X}_{1:n}=\left\{\textbf{x}_1, ..., \textbf{x}_n\right\}$. Consequently, the language generation problem can be defined as finding a function $f$ mapping an input sequence $\textbf{X}$ to a sequence $\textbf{s}$ of $m$ target vectors $\textbf{s}=\textbf{Y}_{1:m}=\left\{\textbf{y}_1, ..., \textbf{y}_m\right\}$. A typical seq2seq model usually consists of three components: (1) An \textbf{encoder}, denoted as $f_{\text{encoder}}$ that accepts an input sequence $\textbf{X}_{1:n}$, and generates a corresponding sequence of contextualized representation $\textbf{h}$. (2) A \textbf{context vector}, $\textbf{c}$ that is a function of $\textbf{h}$ conveying the essence of the input document to the decoder. (3) A \textbf{decoder}, $f_{\text{decoder}}$ that uses $\textbf{c}$ to generate an arbitrary length of sequence $\textbf{Y}_{1:m}$ based on the task-specific requirement.
% \cite{jurafsky2020speech}:

Transformer has become the most effective neural network architecture for natural language modeling \cite{vaswani2017attention}. Comparing to Recurrent Neural Network (RNN), Transformers apply self-attention to compute in parallel every word from the input text an attention weight that gauges the influence each word has on others, thus allowing for parallelization than RNNs for large-scale model training \cite{he2020deberta}. Transformer-based encoder-decoder models are introduced with scaled dot-product attention \cite{vaswani2017attention}. It consists of an encoder and a decoder, both are stacks of residual attention blocks, which can process an input sequence $\textbf{X}_{1:n}$ of variable length $n$ without exhibiting a recurrent structure. This benefits the transformer-based encoder-decoders to be highly parallelizable, which makes the model orders of magnitude more computationally efficient\cite{Platen2020Transformers}.

The transformer-based encoder encodes the input sequence $\textbf{X}_{1:n}$ to a sequence of hidden states $\bar{\textbf{X}}_{1:n}$ (a.k.a. \textbf{h}) by $f_{\text{encoder}}:\textbf{X}_{1:n} \rightarrow \bar{\textbf{X}}_{1:n}$. The transformer-based decoder then models the conditional probability distribution of the target vector sequence $\textbf{Y}_{1:m}$ given the sequence of encoded hidden states $\bar{\textbf{X}}_{1:n}$ as $p_{\text{decoder}}(\textbf{Y}_{1:m}|\bar{\textbf{X}}_{1:n})$. Transformer-based decoder is a stack of decoder blocks followed by a dense layer, the "LM head", which maps the learned sequence of target vectors $\textbf{Y}_{0:i-1}$ to a sequence of logit vectors $\textbf{L}_{1:m}=\left\{\textbf{l}_1,..., \textbf{l}_m\right\}$. Transformer-based decoder defines the conditional probability distribution of a target sequence given the contextualized encoding sequence $p_{\text{decoder}}(\textbf{Y}_{1:m}|\bar{\textbf{X}}_{1:n})$, which by Bayes' rule can be decomposed into a product of conditional distributions as Equation \ref{eq::transformer-decoder}, where $\textbf{W}_{\text{emb}}^T=[\textbf{w}_1, ...,\textbf{w}_{\text{vocab}}]^T$ refers to the transpose of the word embedding matrix, and the logit vector $\textbf{l}_{i}$ represents the similarity score between the encoded output vector and each word embedding. 

\begin{align}
\begin{split}
	P(\textbf{s}|\textbf{d})&=\prod_{i=1}^{m} p_{\text{decoder}}(\textbf{y}_i|\textbf{Y}_{0:i-1}, \bar{\textbf{X}}_{1:n}) \\
	&= \prod_{i=1}^{m} \text{softmax}(f_{\text{decoder}}(\textbf{Y}_{0:i-1}, \bar{\textbf{X}}_{1:n})) \\
	&= \prod_{i=1}^{m} \text{softmax}(\textbf{W}_{\text{emb}}^T \cdot  \text{LMHead}(\bar{\textbf{y}}_{i-1})) \\
	&= \prod_{i=1}^{m} \text{softmax}(\textbf{l}_i) \\
\end{split}
\label{eq::transformer-decoder}
\end{align}

\subsection{Neural Topic Model}
\begin{figure}[htbp]
\centering
\includegraphics[width=0.45\textwidth]{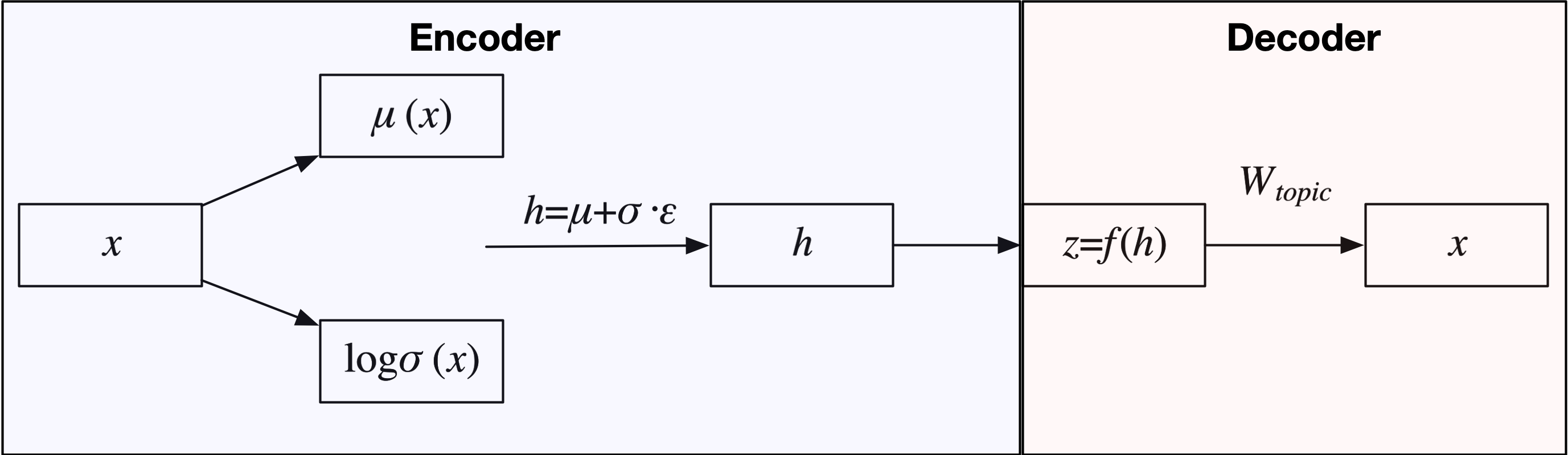}
\caption{Model architecture of Neural Topic Model, as modified from \citet{ding2018coherence}.}
\label{fig::ntm-architecture}
\end{figure}

Topic model aims to extract the implicit topic representation as the short description for a document \cite{blei2003latent}. Neural topic model (NTM) uses neural variational inference \cite{kingma2013auto, rezende2014stochastic} as a flexible framework to accommodate more expressive topic models \cite{ding2018coherence}. Figure \ref{fig::ntm-architecture} is the fundamental model architecture of NTM. $\textbf{x}$ is the representation of the input document and we depress it into a latent topic variable $\textbf{z} \in \mathbb{R}^{K \times 1}$ where $K$ is the number of topics. Usually, \textbf{x} is a bag-of-words representation for an input document for neural topic modeling. We define the structure of NTM following the notations from \citet{ding2018coherence}:

\begin{itemize}[leftmargin=*]
\itemsep0em 
% 	\item In the \textbf{encoder}, we have $\pi=f_{\text{MLP}}(x)$, $\mu(x)=l_1(\pi)$, $\log \sigma(x)=l_2(\pi)$, $h(x, \epsilon)=\mu +\sigma\odot \epsilon$, and finally $z=f(h)$. Different topic models use different definitions for $f$, for example, NTM \cite{miao2016neural} uses ReLU and GSM \cite{miao2017discovering} uses softmax. The function $l_1$ and $l_2$ are linear transformations with bias. We use $q_{\phi}(\textbf{h}|\textbf{x})$ to represent the encoder.
% \item The \textbf{decoder} network $p_{\theta}(\textbf{x}|\textbf{z})$ maps $\textbf{z}$ to the predicted probability of each of the word in the vocabulary $\textbf{y}$ through $\textbf{y} = \text{softmax}(\textbf{W}_{\text{topic}}\textbf{z}+\textbf{b})$, where $\textbf{W}_\text{topic} \in \mathbb{R}^{|V|_{\text{topic}}\times K}$. $|V|_{\text{topic}}$ refers to the vocabulary used for topic model, usually it refers to the top frequent tokens after removing the stopwords. 
    \item In the \textbf{encoder} $q_{\phi}(\textbf{h}|\textbf{x})$, it generates $\mu(\textbf{x})$ and $\sigma(\textbf{x})$ through neural networks to obtain $\textbf{h}=\mu+\sigma\cdot\epsilon$ given the input $\textbf{x}$. Then we learn the latent topic variable $\textbf{z}=f(\textbf{h})$. NTM defines $f$ as ReLU \cite{miao2016neural} and GSM uses softmax \cite{miao2017discovering}.
	\item The \textbf{decoder} network $p_{\theta}(\textbf{x}|\textbf{z})$ maps $\textbf{z}$ to the predicted probability of each word in the vocabulary $\textbf{y}$ through $\textbf{W}_{\text{topic}}$, which usually refers to the top frequent tokens after removing the stopwords. 
% 	The log-likelihood of the document can be written as $\log p_{\theta}(\textbf{x}|\textbf{z})=\sum_{i=1}^{|V|_{\text{topic}}}\left\{x \odot \log y\right\}$.
\end{itemize}

Comparing to the procedure of the language generation of Seq2Seq model in Section \ref{section::transformer-based-seq2seq-model}, we find that the topic modeling task shares several similar features. They both use the encoder-decoder network. The language generation task uses the encoder to understand the input document and generates the task-specific sequence based on the hidden states learned from the encoder. Topic modeling uses the encoder to compress the input document to the latent topic variable, which is considered as the short descriptions of the input document while preserving the essential statistical relationships \cite{blei2003latent}. The latent topic variable is then used to reconstruct the original document via the decoder.

Inspired by the similarity of NTM and Seq2Seq, in this study we incorporate the topic model component into the Seq2Seq model. Similar to \citet{bianchi2020pre, bianchi2020cross}, we use the latent representation from the pre-trained language model as the input for the topic model, which replaces the bag-of-words representation. Here we choose the last hidden state from the encoder as the latent representation for the input document. This contextualized context vector, which refers to the last hidden state in the Transformer-based Seq2Seq model, is expected to carry important information from the input document, on which the decoder relies to generate the output sequence. Using this as the input to the topic model allows us to keep both the valuable symbolic information as BoW and the structure information from the language sequence \cite{bianchi2020cross}. Another benefit of using the context vector is that it can help overcome some limitations from the BoW representation when inferring topics. The objective function of the topic model is to maximize the usual evidence lower bound (ELBO). The loss function of the topic model $\mathcal{L}_{\text{tm}}$ can be written as the negative of the ELBO in Equation \ref{eq::likelihood-topic-model}.
% , where $L$ is the number of words in the document.

\begin{equation}
\begin{split}
	\mathcal{L}_{\text{tm}} \approx D_{\text{KL}}(q_{\phi}(\textbf{h}|\textbf{x})||p_{\theta}(\textbf{h}))-\\ \frac{1}{L}\sum_{l=1}^L \log p_{\theta}(\textbf{x}^i|\textbf{z}^{i,l})
 	\label{eq::likelihood-topic-model}
\end{split}
\end{equation}

\begin{figure*}
\centering
\includegraphics[width=0.8\textwidth]{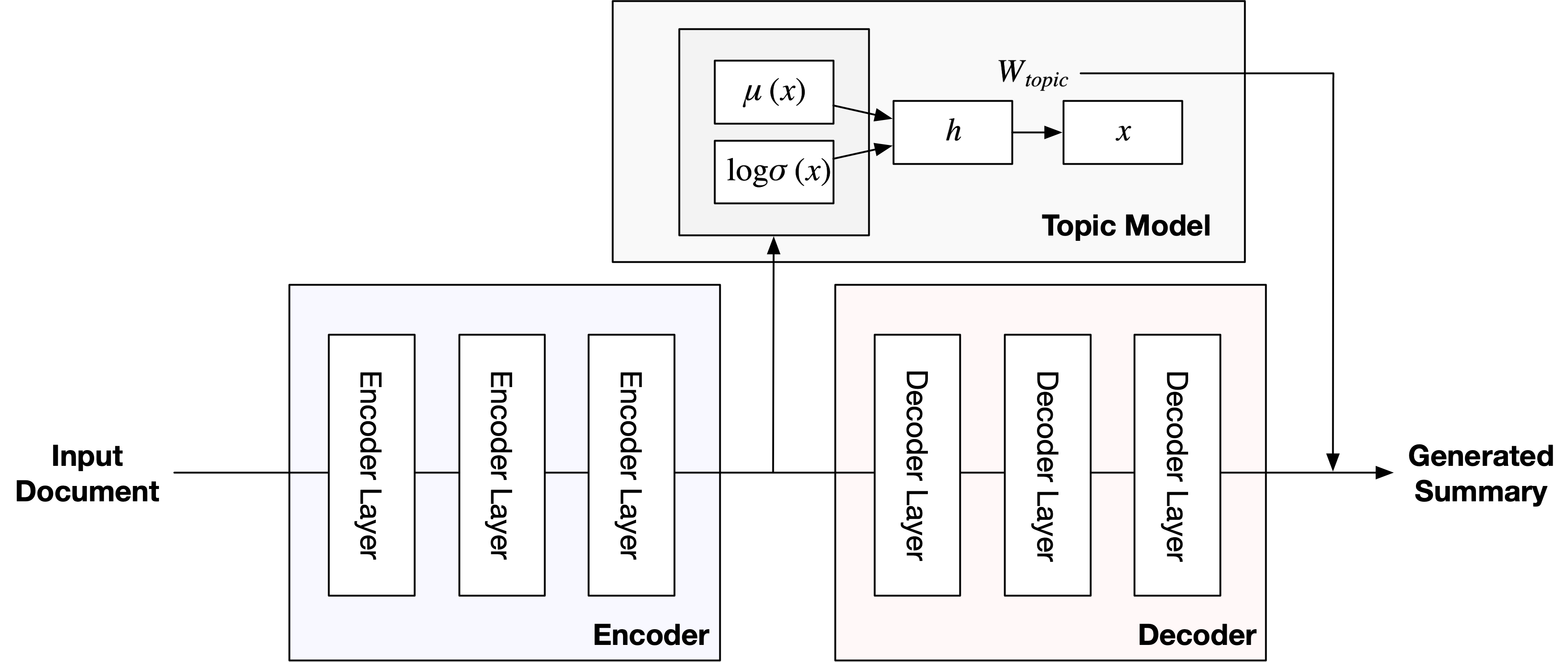}
\caption{Framework of our topic-guided abstractive text summarization model.}
\label{fig::TAS-architecture}
\end{figure*}

\subsection{TAS: Topic-Guided Abstractive Summarization}
Incorporating the topic information into Seq2Seq model, we introduce our model \textbf{TAS}, \underline{T}opic-Guided \underline{A}bstractive \underline{S}ummarization, which learns and extracts the topic-level features to guide the language generation process. The key motivation for this topic-guided generation mechanism is to take the topic distribution as the prior knowledge which can guide the generation module to focus on important source words. To introduce the topic-level features into the language generation, we need to re-define the logit function (see Equation \ref{eq::new-l}). Given a corpus, the topic model learns topics at a global level, which is usually represented as a topic-word matrix (also called topic embedding) $\textbf{W}_{\text{topic}} \in \mathbb{R}^{K \times V_{\text{topic}}}$, where $V_{\text{topic}}$ is the vocabulary size and $K$ is the number of topics. With this topic embedding, we re-write the language generation process as Equation \ref{eq::new-l}. $\text{TMHead}$ is the neural network which transforms the dimension for selecting a word from the vocabulary $\textbf{W}_{\text{emb}}$.

\begin{equation}
\begin{split}
	\textbf{l}_i = \text{softmax}(\textbf{W}_{\text{emb}}^T \cdot  (\text{LMHead}(\bar{\textbf{y}}_{i-1})+\\ \bar{\textbf{y}}_{i-1}\cdot\text{TMHead}(\textbf{W}_{\text{topic}})))
	\label{eq::new-l}
\end{split}
\end{equation}

We expect that the global semantic information can draw enough encoder-decoder attention, which provides guidance on the language generation process. We define the loss function as Equation \ref{eq::loss-function}. The loss for language generation $\mathcal{L}_{\text{finetune}}$ uses the cross-entropy loss. $\alpha$ and $\beta$ are the hyper-parameters for balancing two losses.

\begin{equation}
	\mathcal{L} = \alpha \mathcal{L}_{\text{tm}}+\beta \mathcal{L}_{\text{finetune}}
	\label{eq::loss-function}
\end{equation}

Figure \ref{fig::TAS-architecture} is the framework for our proposed model. Comparing to the prior studies of applying topic-level features to the language generation process (see the discussion in literature \cite{fu2020document}), our model is different in that:
\begin{itemize}[leftmargin=*]
\itemsep0em
	\item We embed the topic model component into the Seq2Seq model, by using the last hidden state from the encoder to infer the topic information.
	\item We incorporate the topic-level features to the language generation process. By building a joint learning framework, our model is able to guide the generation process with the global semantics learned from topics.
\end{itemize}

\section{Experiments}
\subsection{Experimental Setting}
We evaluate our model using two summarization datasets: the CNN/Daily Mail dataset (CNN/DM) \cite{hermann2015teaching} and the extreme summarization dataset (XSUM) \cite{narayan2018don}. Our experiments are conducted with 3 NVIDIA V100 GPUs. We adopt a 12 layer encoder and 6 layer decoder, and our model is warm started with the distil-BART pre-trained model \footnote{We choose distilBART provided by HuggingFace due to fewer parameters. For CNN/DM, we use "sshleifer/distilbart-cnn-12-6". For XSUM, we use "sshleifer/distilbart-xsum-12-6". We record our detailed implementation in Appendix \ref{appendix::implementation}.}. We fine-tune our model for 20 epochs with a batch size of 64. We use Adam optimizer and fine-tune the model with the learning rate of $3e-5$. For the topic model component, we choose the most frequent 2,000 words from the training set as the topic vocabulary for $\textbf{W}_{\text{topic}}$ and set the number of topics to $K=1024$. Our code is available at: https://github.com/chz816/tas.

\subsection{Experimental Results}
We compare our proposed model with the following cutting-edge summarization models, including both extractive and abstractive models.
\begin{itemize}[leftmargin=*]
\itemsep0em
	\item \textbf{Lead-N} uses the first $N$ sentences of the article as its summary.
	\item \textbf{BERTSUM} \cite{liu2019text} proposes a novel document-level encoder based on BERT to generate summary.
	\item \textbf{MATCHSUM} \cite{zhong-etal-2020-extractive} formulates the task as a semantic text matching problem.
	\item \textbf{PGNet} \cite{see2017get} refers to the pointer-generator network. \textbf{PGNet+Cov} is with the coverage mechanism.
	\item \textbf{BART} \cite{lewis2020bart} employs the bidirectional encoder and the left-to-right decoder.
\end{itemize}

We adopt ROUGE \cite{lin2004rouge} F1 score as the evaluation metric. We choose ROUGE-1, ROUGE-2, and ROUGE-L for performance measurement, which are the common choices in the literature. We report the performance of baseline models using the numbers from the original literature.

\textbf{Results on CNN/DM.} Table \ref{table::performance-cnndm} summarizes the evaluation results on the CNN/DM dataset. We compare our model with different types of baselines, including strong extractive and abstractive state-of-the-art models. Our model achieves better performance in most cases. ROUGE-L score of our model is the highest among all models, which emphasizes the success of our model to capture the global semantics on this dataset. Comparing to rule-based baseline and extractive models, TAS achieves a comparable ROUGE-1 score with the extractive SOTA model MATCHSUM and improves the ROUGE-2 and ROUGE-L score by 1.6\% and 1.9\%. TAS also outperforms the pointer-generator network and the abstractive SOTA model BART on CNN/DM. The superior performance of TAS has emphasized the success of introducing the topic model component to help the Seq2Seq model capture the important information from the input document.

\begin{table}[htbp]
\centering
\begin{threeparttable}
\begin{tabular}{lccc}
\toprule
\multicolumn{1}{c}{\textbf{Model}} & \textbf{RG-1} & \textbf{RG-2} & \textbf{RG-L} \\
\hline
Lead-3 & 40.07 & 17.68 & 36.33 \\
\hline
BERTSUM & 42.13 & 19.60 & 39.18 \\
MATCHSUM & \textbf{44.41} & 20.86 & 40.55 \\
\hline
PGNet & 36.44 & 15.66 & 33.42 \\
PGNet+Cov & 39.53  & 17.28 & 36.38 \\
BART & 44.16 & \textbf{21.28} & \textbf{40.90} \\
\hline
TAS & \textbf{44.38} &	\textbf{21.19} & \textbf{41.33}	  \\
\toprule
\end{tabular}
\end{threeparttable}
\caption{ROUGE evaluation on CNN/DM dataset.}
\label{table::performance-cnndm}
\end{table}

\textbf{Results on XSUM.} Table \ref{table::performance-xsum} shows our experimental results on the XSUM dataset. The summary of the XSUM dataset only contains one sentence, which requires the language model to compress the information and generate a precise sentence focusing on the key information. Our proposed model TAS outperforms extractive systems and has achieved comparable performance with BART. 
% Comparing to BART which has a 12-layer encoder and 12-layer decoder, TAS has a smaller architecture with a 12-layer encoder and 6-layer decoder. It is impressive to achieve this comparable performance with less trainable parameters.

\begin{table}[htbp]
\centering
\begin{threeparttable}
\begin{tabular}{lccc}
\toprule
\multicolumn{1}{c}{\textbf{Model}} & \textbf{RG-1} & \textbf{RG-2} & \textbf{RG-L} \\
\hline
Lead-1 & 16.30 & 1.60 & 11.95 \\
\hline
BERTSUM & 38.81 & 16.50 & 31.27 \\
MATCHSUM & 24.86 & 4.66 & 18.41 \\
\hline
PGNet & 29.70 & 9.21 & 23.24 \\
PGNet+Cov & 28.10 & 8.02 & 21.72 \\
BART & \textbf{45.14} & \textbf{22.27} & \textbf{37.25} \\
\hline
TAS & \textbf{44.63} &	\textbf{21.62} & 	\textbf{36.77}  \\
\toprule
\end{tabular}
\end{threeparttable}
\caption{ROUGE evaluation on XSUM dataset.}
\label{table::performance-xsum}
\end{table}

To see whether our model is able to generate meaningful topics as NTM does, we show some topics learned by our model in Appendix \ref{appendix::topics}.

\subsection{Human Evaluation}
We also conduct a human evaluation to further examine the quality of the generated summaries by our model. We choose three metrics which are commonly used in prior studies \cite{huang-etal-2020-knowledge, xu-etal-2020-self}: \textbf{informativeness}, \textbf{fluency} and \textbf{succinctness}. Informativeness measures whether the summary covers the important information from the input article, fluency focuses on if the generated summary is grammatically correct and succinctness measures whether the summary is concise and does not describe too many details. We randomly select 100 articles from the XSUM test set and hire 9 fluent English speakers as our annotators to rate summaries generated by distil-BART\footnote{The reason we choose distil-BART is that TAS is warm started with distil-BART.} and our model TAS. They are required to give a comparison between the two generated summaries that are presented anonymously. Table \ref{table::human-evaluation} reports the human evaluation results. Overall, we find that our model is more capable of capturing the key information of a document and the global semantics. We provide two example summaries generated by our model TAS in Table \ref{table::example-summary}. 
% Source articles are randomly picked from the test set of XSUM. The corresponding generated summaries are unsurprisingly fluent and grammatically correct. 
\begin{table}[htbp]
\centering
\begin{tabular}{l|ccc}
\toprule
 & \textbf{Win} & \textbf{Tie} & \textbf{Loss}\\
 \hline
Informativeness & 35.92\%  & 29.58\% & 34.51\% \\
Fluency & 21.13\% & 60.56\%  & 18.31\% \\
Succinctness & 32.86\% & 41.34\% & 25.80\% \\
\hline
\toprule
\end{tabular}
\caption{Human evaluation results on XSUM dataset. ``Win/Tie/Loss" denotes the comparison of our model with fine-tuned distil-BART.} 
\label{table::human-evaluation}
\end{table}

\begin{table*}
  \centering
  \begin{tabular}{p{260pt}p{160pt}}
  \toprule   
    \multicolumn{1}{c}{\textbf{Source article (abbreviated)}} & \multicolumn{1}{c}{\textbf{Summary by our model}} \\
    \hline
    The 32-year-old scored 25 points on Sunday to pass the milestone in his 10th season with the Cavaliers.He now has twice as many points as Zydrunas Ilgauskas, who is second on the team's all-time list with 10,616.James, who also played for Miami Heat, is eighth on the NBA all-time list with 27,938 career points... & LeBron James has become the second highest scorer in NBA history after passing 10,000 points for the Cleveland Cavaliers.\\
    \hline
    % Researchers found that four out of five children in England who ate school lunches had tried food at school that they had not tried at home. Half of parents questioned said their children had asked for foods they had eaten at school to be cooked at home... & Children who eat school lunches are more likely to try new foods at home, a survey suggests. \\
    % \hline
    The materials, which can sense pressure as sensitively and quickly as human skin,  have been outlined by two groups reporting in Nature Materials.The skins are arrays of small pressure sensors that convert tiny changes in pressure into electrical signals... & Engineers have shown off two approaches to creating flexible "skin" that could be used in robotics and artificial limbs.\\
    % \hline
    % The London trio are up for best UK act and best album, as well as getting two nominations in the best song category."We got told like this morning 'Oh I think you're nominated'", said Dappy."And I was like 'Oh yeah, which one?' And now we've got nominated for four awards. I mean, wow!"... &  N-Dubz have revealed they were surprised to be nominated for four Mobo Awards.\\
    \toprule
  \end{tabular}
\caption{Example summaries by our model on XSUM dataset.} 
\label{table::example-summary}
\end{table*}

\subsection{Model Analysis}
To have a deeper understanding of why our model improves the summarization and the impacts of the topic model component, we conduct several additional experiments using the XSUM test set.

\textbf{Impacts of the topic model.} Topic model is an important component in our model and provides semantic information. We repeat our experiments on XSUM using different types of topic models, including prodLDA \cite{srivastava2017autoencoding} and LDA \cite{blei2003latent}. The main difference between these two models is prodLDA replaces the multinomial distribution over individual words in standard LDA with a product of experts. Table \ref{table::different-type-topics} records the results, and we can clearly see that prodLDA and LDA produce very similar results.

\begin{table}[htbp]
\centering
\begin{tabular}{lccc}
\toprule
\multicolumn{1}{c}{\textbf{Topic Model}} & \textbf{RG-1} & \textbf{RG-2} & \textbf{RG-L} \\
\hline
prodLDA & 44.63 &	21.62 & 	36.77 \\
LDA & 44.63 & 21.60 & 36.78\\
\toprule
\end{tabular}
\caption{Performance for our model with different types of topic models, prodLDA and LDA. We use the same experimental settings with $\alpha=0.1$ and $\beta=1$.}
\label{table::different-type-topics}
\end{table}

\textbf{Impacts of the number of topics.} We fix the number of topics $K=1024$ in our model. To test whether the number of topics affects the model performance, we need to add an additional neural network $\text{TMHead}$ to adjust the dimension for the generation process in Equation \ref{eq::new-l-adjust}. Table \ref{table::different-num-topics} records the performance. We can find that with an additional neural network, the model with $K=1024^*$ topics perform the best and it achieves a similar ROUGE score with our proposed model which fixes the topic number. Considering the better performance of the proposed TAS model and fewer parameters, we report the performance of the model by fixing the topic numbers $K=1024$.

\begin{equation}
\begin{split}
	\textbf{l}_i = \text{softmax}(\textbf{W}_{\text{emb}}^T \cdot  (\text{LMHead}(\bar{\textbf{y}}_{i-1})+\\ \text{DimHead}(\bar{\textbf{y}}_{i-1})\cdot\text{TMHead}(\textbf{W}_{\text{topic}})))
	\label{eq::new-l-adjust}
\end{split}
\end{equation}

\begin{table}[htbp]
\centering
\begin{tabular}{lccc}
\toprule
\multicolumn{1}{c}{\textbf{Topic Number}} & \textbf{RG-1} & \textbf{RG-2} & \textbf{RG-L} \\
\hline
$K=1024$ & 44.63 &	21.62 & 	36.77 \\
\hline
$K=256^*$ & 44.31  & 21.11  & 36.07 \\
$K=512^*$ & 44.20 & 21.05 & 35.98 \\
$K=1024^*$ & 44.55 & 21.56 & 36.73 \\
\toprule
\end{tabular}
\caption{Performance for our model with different numbers of topics. $^*$ indicates the model with additional neural network.} 
\label{table::different-num-topics}
\end{table}

\textbf{Could we just fine-tune the original model in more epochs?} Since our model is warmed up using distil-BART, one could assume that the original distil-BART may simply need to be fine-tuned longer to achieve the same experimental results. Inspired by \citet{peinelt2020tbert}, we perform an additional experiment to finetune distil-BART using the same experimental settings. By analyzing the results in Table \ref{table::finetune-model-analysis}, we can conclude that longer finetuning does not considerably boost distil-BART's performance. This also emphasizes the impacts by introducing the topic model component to our model.

\begin{table}[htbp]
\centering
\begin{subtable}{1\linewidth}\centering
{\begin{tabular}{lccc}
\toprule
\multicolumn{1}{c}{\textbf{Model}} & \textbf{RG-1} & \textbf{RG-2} & \textbf{RG-L} \\
\hline
distil-BART & 44.29  & 21.11	 & 41.18 	 \\
Our model & 44.38 &	21.19 & 	41.33\\
\toprule
\end{tabular}}
\caption{Performance on CNN/DM}
\label{tab:1a}
\end{subtable}
\begin{subtable}{1\linewidth}
\centering
{\begin{tabular}{lccc}
\toprule
\multicolumn{1}{c}{\textbf{Model}} & \textbf{RG-1} & \textbf{RG-2} & \textbf{RG-L} \\
\hline
distil-BART & 44.67  & 21.55	 &  36.62	 \\
Our model & 44.63 &	21.62 & 	36.77\\
\toprule
\end{tabular}}
\caption{Performance on XSUM}
\label{tab:1b}
\end{subtable}
\caption{Performance of distil-BART and our model under the same experimental setting.} 
\label{table::finetune-model-analysis}
\end{table}

We choose one example document (the last example in Table \ref{table::example-summary}), and compare the generated summaries from both fine-tuned distil-BART and our model in Table \ref{tab::generated-summary-TAS-bart}. Our model produces a better summary and is very different from that by distil-BART. Comparing to the target summary, the summary generated by our model accurately captures the key ideas in a detailed manner. Since the only difference between our model and distil-BART is the introduction of topic-level features on generation, we can conclude that the topic model component indeed helps understand the key information and serves as an additional source for details.

\begin{table}[htbp]
\centering
\begin{tabular}{p{40pt}p{150pt}}
\toprule
& \multicolumn{1}{c}{\textbf{Summary}} \\
\hline
Target & "Artificial skin" that could bring a sensitive touch to robots and prosthetic limbs, has been shown off.\\
\hline
Our model & Engineers have shown off two approaches to creating flexible "skin" that could be used in robotics and artificial limbs.\\
\hline
distil-BART & Two approaches to developing flexible "skin" have been presented to scientists.\\
\toprule
\end{tabular}
\caption{Generated summaries from TAS and fine-tuned distil-BART.}
\label{tab::generated-summary-TAS-bart}
\end{table}

\textbf{Ablation study.} We perform ablation studies to test the effects of various hyper-parameters. As shown in Table \ref{table::ablation}, we report the performance under different values of $\alpha$. We fix $\beta=1$ and find that our model achieves the best performance when $\alpha=0.1$. In general, we find that different parameters, which balance the loss between two components, provide different performance, and smaller values on $\alpha$ usually perform better.

\begin{table}[htbp]
\centering
\begin{tabular}{lccc}
\toprule
\multicolumn{1}{c}{$\alpha$} & \textbf{RG-1} & \textbf{RG-2} & \textbf{RG-L} \\
\hline
0 & 44.57 & 21.60 & 36.71	 \\
0.1 & \textbf{44.63} & \textbf{21.62} & \textbf{36.77}\\
1 & 44.28 & 21.06 & 35.97 \\
\toprule
\end{tabular}
\caption{Ablation studies on XSUM dataset, using different values for $\alpha$. We fix $\beta=1$ and use prodLDA for the topic model component.} 
\label{table::ablation}
\end{table}

\subsection{Robustness Check}
In this section, we perform a robustness check for the proposed model. We use the heuristics used in \citet{goel2021robustness} to create sub-populations and perform experiments to check the performance of these sub-populations. We use the following metrics and first select the top 10\% and bottom 10\% examples in the test set of XSUM as two sub-populations.
\begin{itemize}[leftmargin=*]
\itemsep0em
    \item \textbf{Length} of the input document.
    \item \textbf{Position} is the average location in the source document of where information in the summary comes from.
\end{itemize}

Table \ref{table::robustness} records the experimental results, using ROUGE-2 for evaluation \footnote{We report ROUGE-1, ROUGE-2 and ROUGE-L score for robustness check in Appendix \ref{appendix::robustness-check}.}. We find that TAS is more capable to handle different lengths of documents since the performance on the shortest and longest set is better. The design of TAS is aimed to add the topic model component to provide more topic-level information to guide the summary generation. The performance improvement on the longest document set especially encourages us, since we are motivated to find that TAS performs better when the input document is long which is harder to summarize. For position, we find that TAS performs better than the baseline in the latest position set, which is challenging. The latest position group requires our model to have a better understanding of the whole document. TAS achieves a higher ROUGE-2 score in this subpopulation than our baseline by 1.4\%.

\begin{table}[htbp]
\centering
\begin{tabular}{llcc}
\toprule
\multicolumn{2}{c}{\textbf{Metric}} & \textbf{Baseline} & \textbf{TAS}\\
\hline
\multirow{2}{*}{Length} & Shortest & 23.55 & 23.71 \\
& Longest & 15.48 & 15.70 \\
\hline
\multirow{2}{*}{Position} & Earliest & 22.50 & 22.52 \\
& Latest & 17.37 & 17.61 \\
\toprule
\end{tabular}
\caption{Robustness Check on sub-populations defined by metrics. For Baseline model, we use the results from distil-BART. We report the performance using ROUGE-2 score.} 
\label{table::robustness}
\end{table}

\section{Conclusion}
In this paper, we propose to leverage the topic model incorporated into the Seq2Seq model to improve the performance of text summarization. We introduce the global semantic topic-level features into the language understanding to guide the generation process. Experiments on two datasets show that our model outperforms several strong baselines in both quantitative measurement and human evaluation, which demonstrates the effectiveness of our proposed model.

\clearpage
\bibliography{anthology,custom}
\bibliographystyle{acl_natbib}

\clearpage
\appendix

\section{Implementation}
\label{appendix::implementation}
% Our proposed model TAS in this paper is warmed up using the pre-trained model from HuggingFace: for CNN/DM, we use "sshleifer/distilbart-cnn-12-6" \footnote{https://huggingface.co/sshleifer/distilbart-cnn-12-6}. For XSUM: we use "sshleifer/distilbart-xsum-12-6" \footnote{https://huggingface.co/sshleifer/distilbart-xsum-12-6}. 
% These two models are fine-tuned on the corresponding datasets. Comparing to the original BART, distil-BART contains much less parameters, which is also able to achieve comparable performance. 
For CNN/DM, we also add the post-processing step after the generation. We use the same post-processing program \footnote{https://github.com/microsoft/ProphetNet} from SOTA model ProphetNet \cite{qi2020prophetnet}. We don't perform post-processing for XSUM, considering the summary for XSUM only contains one sentence.

\section{Learned Topics}
\label{appendix::topics}
We show some learned topics in this section to check if the topic model component can generate meaningful topics as expected. Table \ref{table::example-topics} lists the example topics learned on CNN/DM and XSUM dataset. For each topic, we extract the top 10 words from the topic model vocabulary $\textbf{W}_{\text{topic}}$. By analyzing the learned topics, we find that our topic model component has learned several topics which are explainable and meaningful.

\begin{table}[htbp]
\centering
\begin{subtable}{1\linewidth}\centering
{\begin{tabular}{p{5pt}p{185pt}}
\toprule
& \multicolumn{1}{c}{\textbf{Top 10 words}} \\
\hline
1 & court, said, police, told, case, death, attorney, mr, judge, charges\\
2 & family, told, life, mother, said, children, old, like, time, would\\
3 & government, united, cnn, military, security, states, officials, international, al, forces\\
4 & season, league, team, game, cup, win, match, club, goal, players\\
5 & water, according, scientists, could, agency, earth, officials, government, per, region\\
\toprule
\end{tabular}}
\caption{Example topics learned on CNN/DM.}
\label{table::example-topics-cnndm}
\end{subtable}
\begin{subtable}{1\linewidth}
\centering
{\begin{tabular}{p{5pt}p{185pt}}
\toprule
& \multicolumn{1}{c}{\textbf{Top 10 words}} \\
\hline
1 & league, game, season, first, half, team, second, win, side, goal\\
2 & police, family, court, found, old, heard, hospital, told, incident, died\\
3 & health, service, site, work, company, use, help, children, patients, customers\\
4 & first, game, league, win, season, team, club, side, second, players\\
5 & mr, government, president, us, state, said, minister, security, country, un\\
\toprule
\end{tabular}}
\caption{Example topics learned on XSUM.}
\label{table::example-topics-xsum}
\end{subtable}
\caption{Example topics learned by our model.} 
\label{table::example-topics}
\end{table}

\section{Robustness check}
\label{appendix::robustness-check}
We report the ROUGE-1, ROUGE-2 and ROUGE-L score for robustness check to better understand our proposed model in Table \ref{table::robustness-apendix}.

\begin{table}[htbp]
\centering
\begin{subtable}{1\linewidth}\centering
{\begin{tabular}{llcc}
\toprule
\multicolumn{2}{c}{\textbf{Metric}} & \textbf{Baseline} & \textbf{TAS}\\
\hline
\multirow{2}{*}{Length} & Shortest & 45.84 & 45.84 \\
& Longest & 36.92 & 36.98 \\
\hline
\multirow{2}{*}{Position} & Earliest & 45.35 & 45.31 \\
& Latest & 39.46 &  39.81\\
\toprule
\end{tabular}}
\caption{Performance evaluated by ROUGE-1.}
\label{table::rouge-1}
\end{subtable}
\begin{subtable}{1\linewidth}
\centering
{\begin{tabular}{llcc}
\toprule
\multicolumn{2}{c}{\textbf{Metric}} & \textbf{Baseline} & \textbf{TAS}\\
\hline
\multirow{2}{*}{Length} & Shortest & 23.55 & 23.71 \\
& Longest & 15.48 & 15.70 \\
\hline
\multirow{2}{*}{Position} & Earliest & 22.50 & 22.52 \\
& Latest & 17.37 & 17.61 \\
\toprule
\end{tabular}}
\caption{Performance evaluated by ROUGE-2.}
\label{table::rouge-2}
\end{subtable}
\begin{subtable}{1\linewidth}
\centering
{\begin{tabular}{llcc}
\toprule
\multicolumn{2}{c}{\textbf{Metric}} & \textbf{Baseline} & \textbf{TAS}\\
\hline
\multirow{2}{*}{Length} & Shortest & 39.55 & 39.72 \\
& Longest &  28.84& 29.09 \\
\hline
\multirow{2}{*}{Position} & Earliest & 37.72 & 37.8  \\
& Latest & 31.51 & 31.81 \\
\toprule
\end{tabular}}
\caption{Performance evaluated by ROUGE-L.}
\label{table::rouge-l}
\end{subtable}
\caption{Robustness Check with ROUGE-1, ROUGE-2 and ROUGE-L.} 
\label{table::robustness-apendix}
\end{table}

\end{document}